\documentclass[letterpaper, 10 pt, conference]{ieeeconf}  

\IEEEoverridecommandlockouts                              

\usepackage{graphics} 
\usepackage{epsfig} 
\usepackage{mathptmx} 
\usepackage{times} 
\usepackage{amsmath} 
\usepackage{amssymb}  
\usepackage{bm}

\usepackage{graphicx}
\usepackage{color}
\usepackage{import}
\usepackage[font=footnotesize]{caption} 
\usepackage{subcaption}
\usepackage{hyperref}
\usepackage{cite}
\usepackage{float}
\usepackage[percent]{overpic} 

\title{\LARGE \bf
Bootstrapped Self-Supervised Training with Monocular Video for Semantic Segmentation and Depth Estimation
}

\author{Yihao Zhang and John J. Leonard
\thanks{Yihao Zhang and John Leonard are with the Computer Science and Artificial Intelligence Laboratory, Massachusetts Institute of Technology, Cambridge, MA 02139 USA {\tt\small \{yihaozh, jleonard\}@mit.edu}}%
\thanks{This work was supported by ONR MURI grant N00014-19-1-2571,
ONR grant N00014-18-1-2832, and ARPA-E award
DE-AR0001218 under the DIFFERENTIATE Program.}}

\begin{document}

\maketitle
\thispagestyle{empty}
\pagestyle{empty}

\begin{abstract}

For a robot deployed in the world, it is desirable to have the ability of autonomous learning to improve its initial pre-set knowledge. We formalize this as a bootstrapped self-supervised learning problem where a system is initially bootstrapped with supervised training on a labeled dataset and we look for a self-supervised training method that can subsequently improve the system over the supervised training baseline using only unlabeled data. In this work, we leverage temporal consistency between frames in monocular video to perform this bootstrapped self-supervised training. We show that a well-trained state-of-the-art semantic segmentation network can be further improved through our method. In addition, we show that the bootstrapped self-supervised training framework can help a network learn depth estimation better than pure supervised training or self-supervised training.

\end{abstract}

\section{Introduction}

We consider a practical problem in robotics, bootstrapped self-supervised learning, where a system is initially trained on a labeled dataset and later on this bootstrapped system seeks to improve itself through self-supervised training with unlabeled data. This unique problem lies at the intersection of two areas in machine learning; it is a more restricted case of semi-supervised learning \cite{van2020survey} in that unlabeled data are used after labeled data but not in the reverse order nor jointly; unsupervised domain adaptation is a special case of bootstrapped self-supervised learning when the unlabeled data come from a different domain than the labeled data. Due to its uniqueness, bootstrapped self-supervised learning is rarely explored in the machine learning literature even though such a capability would be quite useful in robotics, with the aim of long-term autonomy.

In machine learning, the most common way of using unlabeled data is through unsupervised pre-training on proxy tasks to learn common feature representations before any supervised fine-tuning with labeled data \cite{zhan2018mix, wang2015unsupervised, agrawal2015learning, pathak2017learning, girshick2014rich, grill2020bootstrap}. Bootstrapped self-supervised training is exactly the opposite of this process (i.e. labeled data training first followed by unlabeled data training). The absence of ground truth labeling in the final training stage makes the problem particularly challenging.

In this work, we develop a method by leveraging temporal consistency available in video to realize bootstrapped self-supervised training in the cases of learning single-image semantic segmentation and depth estimation. For semantic segmentation, we show that through our additional self-supervised training, the performance of a state-of-the-art network can be enhanced over the supervised training baseline both when the supervised training dataset is the same as the self-supervised training dataset and when they are different. For depth estimation, we demonstrate that the bootstrapped self-supervised training strategy is better than pure supervised training or self-supervised training. 

\setlength{\belowcaptionskip}{-10pt}
\setlength{\abovecaptionskip}{2.0pt}
\begin{figure}[t!]
    \vspace{5pt} 
    \centering
    \begin{subfigure}{0.48\columnwidth}
        \includegraphics[width=1.0\textwidth]{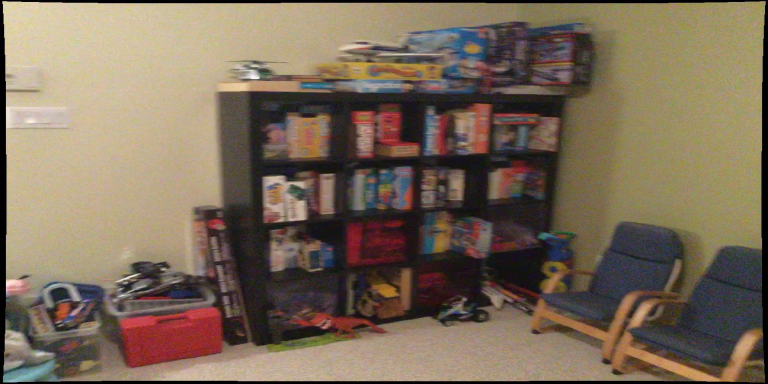}
    \end{subfigure}
    \begin{subfigure}{0.48\columnwidth}
        \includegraphics[width=1.0\textwidth]{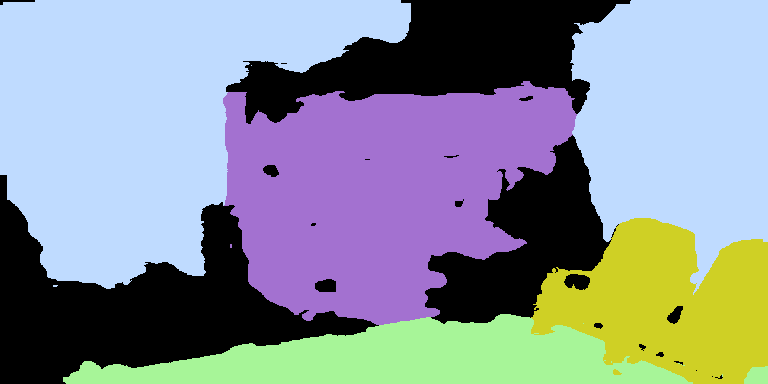}
    \end{subfigure}
    \par\smallskip
    \begin{subfigure}{0.48\columnwidth}
        \includegraphics[width=1.0\textwidth]{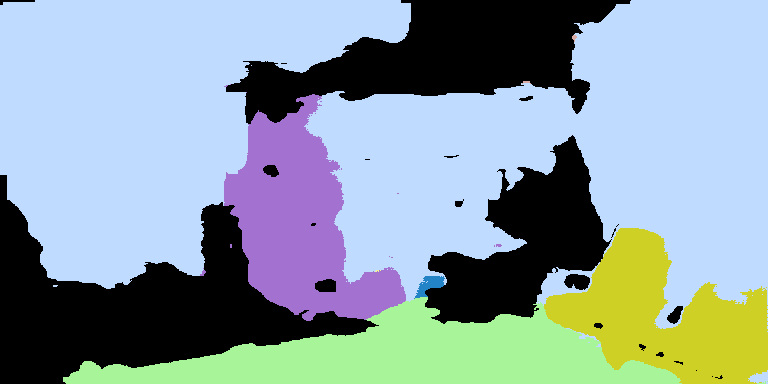}
    \end{subfigure}
    \begin{subfigure}{0.48\columnwidth}
        \includegraphics[width=1.0\textwidth]{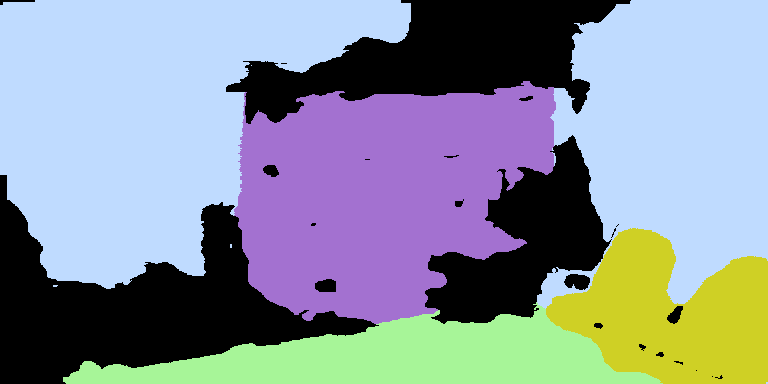}
    \end{subfigure}
    \caption{An example comparison between the semantic segmentation prediction from a supervised training baseline (lower left) and the one after the bootstrapped self-supervised training (lower right) for an input image and its ground-truth labeling (top row) from the ScanNet data \cite{dai2017scannet}.}
    \label{fig: intro}
\end{figure}

\section{Related Work}
\label{sect: literature}

Our work focuses on utilizing the temporal consistency existing in unlabeled monocular video to improve the network performance over a supervised training baseline. We review the prior literature on how temporal consistency in video can help learn semantic segmentation and depth estimation.
\subsection{Learning Semantic Segmentation with Video} 
The use of temporal consistency in video to improve semantic segmentation can be dated back to when conditional random field (CRF) was the dominant method for semantic segmentation. Floros and Leibe \cite{floros2012joint} applied a temporal consistency potential in a CRF to enforce that the same label is given to those pixels that correspond to the same 3D point. They found the correspondences by building a 3D point cloud using stereo visual odometry and mapping. Their idea of temporal consistency of semantic labels is essentially similar to ours, but they realized it in the CRF framework instead of the modern CNN-based end-to-end training framework. 

Later CNN approaches considered temporal consistency through different avenues. One common direction is to take advantage of the pre-training and fine-tuning paradigm \cite{wang2015unsupervised, agrawal2015learning, pathak2017learning}. These methods typically pre-train the network on a video-based proxy task, such as motion segmentation \cite{pathak2017learning}, camera ego-motion prediction \cite{agrawal2015learning}, and feature map matching for video image patches \cite{wang2015unsupervised}, to learn common underlying feature representations. After the pre-training, they fine-tune their networks with labeled data on the target task.

Another line of work addresses temporal consistency by fusing feature maps internally in the network across time \cite{gadde2017semantic, pfeuffer2019semantic, paul2020efficient}. They typically warp the feature maps of a past frame to the current frame and fuse the warped feature maps with those of the current frame. This line of work is often called video semantic segmentation as the networks have to be trained with semantically segmented video and run on video. Unlike these methods, our method uses unlabeled video for the self-supervised training phase and is not restricted to operate on video at run-time. 

The last set of methods utilizes temporal consistency to create labeled images from unlabeled ones \cite{tokmakov2016weakly, zhu2019improving, chen2020naive}. Tokmakov et al. \cite{tokmakov2016weakly} generated pixel labels automatically from image-level labels using video object segmentation. Zhu et al. \cite{zhu2019improving} propagated ground-truth pixel labels to nearby frames with video prediction techniques to create artificially labeled images, and thereby augmented the size of the training set. Our idea is similar to Zhu et al.'s in the sense that we both consider the pixel-level temporal consistency of labels. 

The unsupervised pre-training approaches and the pseudo-labeling approaches mentioned above all belong to semi-supervised learning \cite{van2020survey}. However, they either use unlabeled data before labeled data \cite{wang2015unsupervised, agrawal2015learning, pathak2017learning} or jointly \cite{tokmakov2016weakly, zhu2019improving, chen2020naive}, so the network still sees ground truth labeled data in the last training stage. Whereas in our scenario of bootstrapped self-supervised training, the network does not see labeled data in the final self-supervised training stage, which is a more challenging case. Methodologically, the pre-training approaches try to learn common feature representations from unlabeled data and the pseudo-labeling approaches augment the training data size by transferring labels to unlabeled images, while we try to use unlabeled data to spot deficiencies (in our case temporal inconsistencies) in the prediction and create a training signal to correct the deficiencies.
\subsection{Self-Supervised Learning of Depth and Ego-Motion}
Prior work on self-supervised learning of depth and ego-motion \cite{zhou2017unsupervised, mahjourian2018unsupervised, wang2018learning, godard2019digging, bian2019unsupervised} has explored temporal photometric consistency which states that if a pixel in one frame and a pixel in a second frame both correspond to the same 3D point, they should have the same intensity value in the images. As the correspondence can be set up by knowing the depth at one of the pixels and the ego-motion of the camera between the two frames, minimizing the photometric difference naturally becomes an objective function for training the depth and ego-motion predictions with monocular video. 

Our method extends the prior work on self-supervised learning of depth and ego-motion \cite{zhou2017unsupervised, mahjourian2018unsupervised}. The key difference is that we do not only have a depth network and an ego-motion network but also a semantic segmentation network. We also take advantage of the supervised training results in the self-supervised training stage.

\section{Method}

\subsection{Networks}
\label{sect: nets}
Three networks are used in our approach. The depth network which is a DispNet \cite{mayer2016large} takes in an RGB frame $I_t$ and predicts its depth map $\hat{D}_t$ where $t$ denotes a time instance. The semantic segmentation network, which has the AdapNet++ architecture \cite{valada2019self}, takes in an RGB frame $I_t$ and predicts a semantic segmentation of the frame $\hat{S}_t$ in the form of softmax probabilities. We choose AdapNet++ because of its public availability and its superior performance on several benchmarks \cite{Cordts2016Cityscapes, dai2017scannet, song2015sun}. The third network is an ego-motion network \cite{zhou2017unsupervised} which takes in a set of consecutive frames and outputs the relative $SE(3)$ camera poses between the frames $\hat{T}_{t \rightarrow t'}$ where $t'$ denotes a time instance near time $t$. We also load the depth network with the task of predicting an outlier mask $\hat{O}_t$ (called explainability mask in \cite{zhou2017unsupervised}) which takes a value within $[0, 1]$ at each pixel to cover outlier pixels in the photometric consistency loss due to dynamic objects and static occlusions. The depth network is modified to branch out at the last four feature map levels to enable this prediction. 

\subsection{Supervised Training}
In the bootstrapped self-supervised training scheme, the depth network and the semantic segmentation network are first trained in a supervised manner. The training follows the standard protocol of using cross-entropy loss to learn semantic segmentation and L1 loss to learn depth estimation. The losses are enforced on the outputs at four different scale levels in the DispNet architecture (as in \cite{zhou2017unsupervised}) and three different scale levels in the AdapNet++ architecture (as in \cite{valada2019self}). We perform random scaling and cropping on the input RGB images for data augmentation. Any other advanced supervised training techniques (e.g. pre-training on ImageNet \cite{deng2009imagenet}) are welcome. In our experiments, we directly use the publicly available AdapNet++ checkpoints to reflect the state-of-the-art supervised training baseline. The ego-motion prediction and the outlier mask prediction are left unattended in this step since we consider a practical setting and the ground truth of these two is more difficult to obtain in practice (e.g. it is difficult to accurately measure the trajectory of a hand-held camera).

\subsection{Bootstrapped Self-Supervised Training}
\label{sect: bsst}

\begin{figure*}[t]
    \centering
    \vspace{7pt}
    \begin{overpic}[width=0.85\textwidth, tics=5]{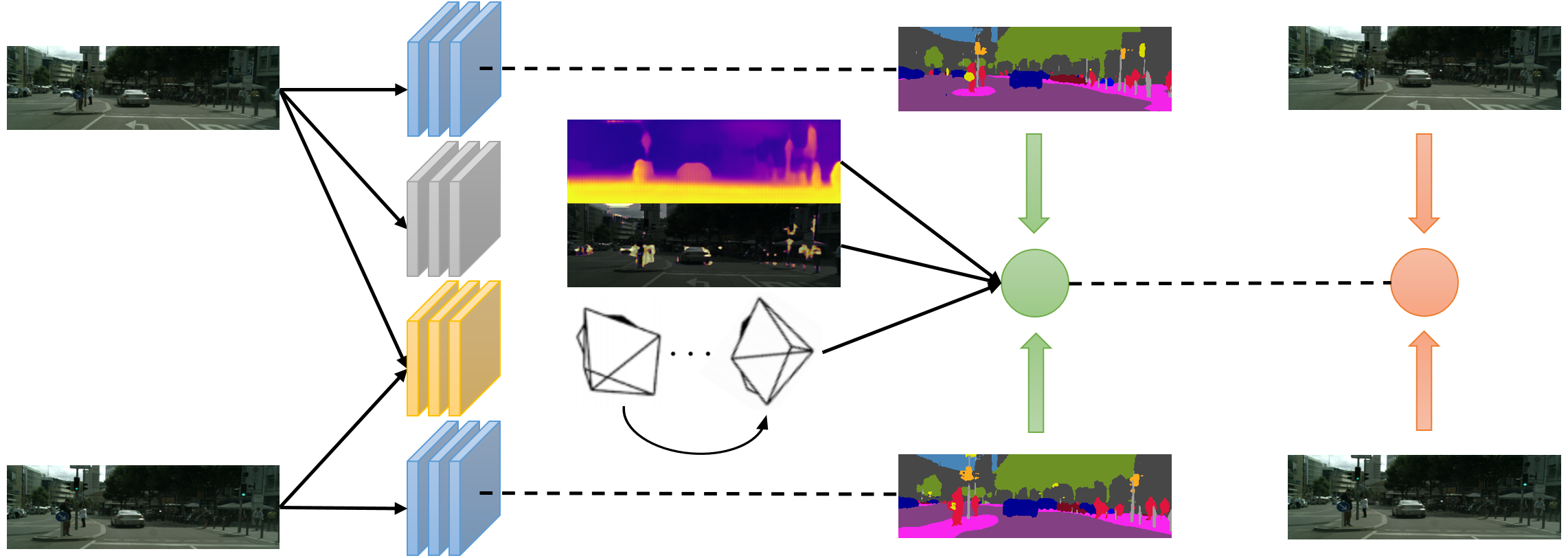}
        \put(42.5,8){$\hat{T}_{t \rightarrow t'}$} \put(54.7,32.2){$\hat{S}_t$} \put(54.7,5.7){$\hat{S}_{t'}$}
        \put(9,24.5){$I_t$} \put(9,7.3){$I_{t'}$} \put(33,23){$\hat{D}_t$} \put(33,18){$\hat{O}_t$}
        \put(68.5,19){$L_{sc}$}
        \put(93.5,19){$L_{pho}$} \put(93.5,14.5){$L_{ssim}$}
    \end{overpic}
    \caption{An illustration of the scheme to compute the consistency losses. Blue network: semantic segmentation network. Gray network: depth and outlier mask network. Yellow network: ego-motion network. Green circle: semantic consistency loss. Orange circle: photometric consistency loss and SSIM loss.} 
    \label{fig: illu}
\end{figure*}
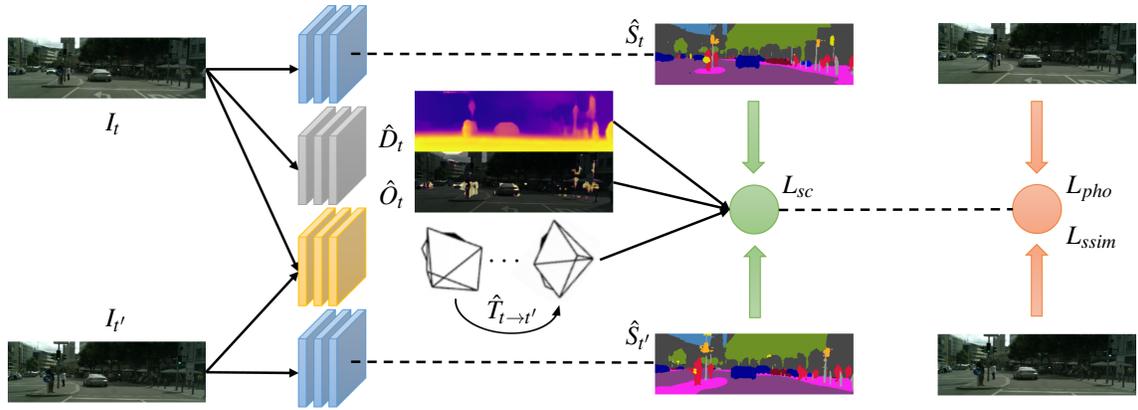

In the bootstrapped self-supervised training stage that follows the supervised training, the networks are trained on unlabeled monocular video. We divide the losses in this training phase into two groups; the temporal consistency losses are the key to improving the prediction quality, whereas the prior losses impose what we already know or have learned. The meaning of bootstrapping is two-fold. First, the network parameters in the self-supervised training phase are initialized by the parameters learned in the supervised training. Second, the supervised training predictions are applied into the prior losses.
\newline

\noindent{\bf Consistency losses.} To enforce the temporal consistency in the end-to-end training with monocular video, it is first necessary to find the corresponding pixels given two nearby frames. This is done by perspective projection:
\begin{equation}
    p' \sim K\hat{T}_{t \rightarrow t'}\hat{D}_t(p)K^{-1}p
\label{eq: projection}
\end{equation}
where $\hat{T}_{t \rightarrow t'}$ and $\hat{D}_t$ are defined in Section \ref{sect: nets}, $p$ denotes the homogeneous coordinates of a pixel in frame $I_t$, $p'$ is the projected pixel in frame $I_{t'}$ and $K$ is the camera intrinsics matrix. The projected coordinates $p'$ do not hold integer values for indexing pixels. Thus, to compute the image intensity value at $p'$, differentiable bilinear interpolation is used \cite{jaderberg2015spatial}.

Now that we have the corresponding pixels, we can write the photometric consistency loss \cite{zhou2017unsupervised} between the two given frames as
\begin{equation}
    L_{pho} = \frac{1}{|V|}\sum_{p\in{V}}{\hat{O}_t(p)\big|I_{t'}(p') - I_t(p)\big|}
\label{eq: photoloss}
\end{equation}
where $V$ is the set of valid pixels projected within the image borders of frame $I_{t'}$ \cite{bian2019unsupervised}. This loss encodes the idea that the corresponding pixels in the two frames should have the same intensity value. This consistency creates a training signal to the depth network and the ego-motion network through $p'$. The outlier mask $\hat{O}_t(p)$ is automatically learned during the self-supervised training to down-weight the loss at pixel $p$ when the residual is too large at pixel $p$ possibly because pixel $p$ breaks the photometric consistency. This often happens when pixel $p$ belongs to a dynamic object or the corresponding 3D point of pixel $p$ is occluded in frame $I_{t'}$. As in \cite{mahjourian2018unsupervised}, we also incorporate a structured similarity (SSIM) loss \cite{wang2004image} to further encourage the similarity between the image patches around $p$ and $p'$: 
\begin{equation}
    L_{SSIM} = \frac{1}{|V|}\sum_{p\in{V}}{\hat{O}_t(p)\big[1 - SSIM(p',p)\big]}
\label{eq: ssimloss}
\end{equation}
where $SSIM(p',p) = \frac{[2\mu(p)\mu(p')+c_1][2\sigma(p,p')+c_2]}{[\mu^2(p)+\mu^2(p')+c_1][\sigma(p)+\sigma(p')+c_2]}$ where $\mu(p)$ and $\sigma(p)$ are the local mean and variance of a $3 \times 3$ patch around pixel $p$, $\sigma(p,p')$ is the covariance between the patches at pixel $p$ and $p'$, $c_1 = 0.01^2$, and $c_2 = 0.03^2$. Equations (\ref{eq: photoloss}) and (\ref{eq: ssimloss}) are for training the the depth and ego-motion networks. We additionally include a semantic consistency loss to train the semantic segmentation network:
\begin{equation}
    L_{sc} = \frac{1}{|V|}\sum_{p\in{V}}{\hat{O}_t(p)\big|\hat{S}_{t'}(p') - \hat{S}_t(p)\big|}
\label{eq: smticloss}
\end{equation}
A similar loss was used in \cite{zhi2019scenecode} to improve semantic segmentation online in a SLAM system. We experiment the use of this loss in the self-supervised training of a network. Since $\hat{S}(p)$ is a vector of softmax probabilities, a sophisticated distance metric on probabilities can be used. However, similar to \cite{zhi2019scenecode} where the Euclidean distance was applied, we simply use the L1 distance. During training, to prevent the depth and ego-motion networks from being trained on predicted labels and potentially being corrupted by wrong predicted labels, we do not propagate the gradients from the semantic consistency loss (\ref{eq: smticloss}) to the depth and ego-motion networks.

The underlying idea of loss (\ref{eq: smticloss}) is that the semantic segmentation learned in the supervised training step may not have equal quality for views of the same scene at different angles and distances. The consistency loss trains the network to push the segmentations of different views into an agreement. The overall bootstrapped self-supervised training scheme including the consistency losses is shown in Figure \ref{fig: illu}. 
\newline

\noindent {\bf Prior losses.} The prior losses encode what we already know and oftentimes also act as regularization to avoid trivial solutions. The most common prior loss on the depth prediction is the edge-preserving depth smoothness loss \cite{mahjourian2018unsupervised}:
\begin{equation}
    L_{sm} = \sum_{p}{\big|\partial_x \underline{\hat{D}}(p)\big|e^{-|\partial_x I(p)|} + \big|\partial_y \underline{\hat{D}}(p)\big|e^{-|\partial_y I(p)|}}
\label{eq: dsmth}
\end{equation}
where the subscript $t$ is dropped here and hereafter for cleaner notations, and $\underline{\hat{D}} = \hat{D}/\overline{\hat{D}}$ is the predicted depth map divided by the average predicted depth. The is used to make the depth smoothness loss scale-invariant as the photometric loss (\ref{eq: photoloss}) is scale-invariant \cite{wang2018learning}.

Another prior loss is a cross-entropy loss on the outlier mask with an all-ones mask (i.e.\ all inliers) \cite{zhou2017unsupervised}, which is simply
\begin{equation}
    L_{om} = -\sum_{p}{\log{\hat{O}(p)}}
\label{eq: omreg}
\end{equation}
This encodes our understanding that most pixels are inliers and obey the photometric consistency. It is also a regularization loss without which the outlier mask would degrade to zero everywhere during the self-supervised training.

The next two prior losses are for the bootstrapping purpose:
\begin{gather}
    L_D = \sum_p{\big|\hat{D}(p) - \hat{D}_{pre}(p)\big|}
    \label{eq: depthprior} \\
    L_S = \sum_p{\big|\hat{S}(p) - \hat{S}_{pre}(p)\big|}
    \label{eq: smticprior}
\end{gather}
where $\hat{D}_{pre}$ is the prediction made by a copy of the depth network frozen at the end of the supervised training phase, and similarly for $\hat{S}_{pre}$. These two prior losses enable the pre-trained predictions to continue influencing the self-supervised training phase. In this way, we do not discard what we have gained in the supervised pre-training. In addition, the semantic prior loss (\ref{eq: smticprior}) also behaves as regularization to prevent the trivial but temporally consistent solution of all pixels taking the same label.
\newline

\noindent {\bf Remark.} Compared to self-supervised learning of depth and ego-motion, our method introduces one more task of semantic segmentation. On the other hand, compared to pure self-supervised methods, our method uses supervised training for bootstrapping, which means not only initializing the network weights with the pre-trained weights from the supervised training but also applying the supervised training results into the prior losses (\ref{eq: depthprior}) (\ref{eq: smticprior}). This set-up of performing supervised training first and self-supervised training in the next is both practical in the robotics context and superior to supervised training only or self-supervised training only as will be shown in Section \ref{sect: results}.

\subsection{Training and Implementation} 
Although the temporal consistency losses (\ref{eq: photoloss}), (\ref{eq: ssimloss}), and (\ref{eq: smticloss}) were illustrated with a pair of consecutive frames $I_t$ and $I_{t'}$, the networks are trained on 3-frame snippets as in \cite{mahjourian2018unsupervised}. We also experimented using 5-frame snippets but the results were poorer. Each time the two pairs of consecutive frames in a 3-frame snippet, $(I_{t_1}, I_{t_2})$ and $(I_{t_2}, I_{t_3})$, are fed into the networks, and the two frames in a pair both play once the roles of $I_t$ and $I_{t'}$ to train on both forward and backward camera motions. In addition, following \cite{zhou2017unsupervised}, the losses are enforced on the network outputs at multiple scales $\{s\}$ (four scale levels for the depth network and three scale levels for the semantic segmentation network). Therefore, the total loss at each iteration is
\begin{gather}
    L = \sum_{s}\sum_{t\in{\{t_1,t_2,t_3\}}}{L^s_{prior,t}}\sum_{t'\in{n(t)}}{L^s_{t \rightarrow t'}}
\end{gather}
where $L^s_{prior,t}$ is a weighted sum of the prior losses (\ref{eq: dsmth}), (\ref{eq: omreg}), (\ref{eq: depthprior}), (\ref{eq: smticprior}) at scale $s$ for the frame at time $t$, $L^s_{t \rightarrow t'}$ is a weighted sum of the temporal consistency losses (\ref{eq: photoloss}), (\ref{eq: ssimloss}), (\ref{eq: smticloss}) at scale $s$, and $n(t)$ stands for the set of consecutive time instance(s) around $t$ in a 3-frame snippet (e.g. $n(t_1)=\{t_2\}$, $n(t_2)=\{t_1,t_3\}$).

The weights for the losses were only lightly tuned in our experiments. We used the same set of weights for the consistency losses (\ref{eq: photoloss}) (\ref{eq: ssimloss}) (\ref{eq: smticloss}) and the semantic prior loss (\ref{eq: smticprior}). For the depth smoothness loss (\ref{eq: dsmth}), depth prior loss (\ref{eq: depthprior}), and outlier mask regularization (\ref{eq: omreg}), we adapted the weights to the scene types (outdoor vs. indoor) and whether it is a cross-dataset experiment. We observe that when the supervised training dataset and the self-supervised training dataset are different, a smaller prior loss weight can be used. The weights for the losses in the order of the equations (\ref{eq: photoloss})--(\ref{eq: smticprior}) are 1.0, 0.15, 0.8, 0.025, 0.08, 0.08, 1.5 (Section \ref{sect: sn2sn}), 1.0, 0.15, 0.8, 0.01, 0.07, 0.03, 1.5 (Section \ref{sect: sun2sn}), 1.0, 0.15, 0.8, 0.07, 0.08, 0.08, 1.5 (Section \ref{sect: cs2cs}), and 1.0, 0.15, 0.8, 0.01, 0.07, 0.03, 1.5 (Section \ref{sect: cs2k}).

The implementation of the bootstrapped self-supervised training framework is based on the public implementation of \cite{mahjourian2018unsupervised} which is for learning depth and ego-motion predictions from video. The ego-motion network is unmodified, while the depth network is modified according to Section \ref{sect: nets} to allow the outlier mask prediction. We add a few new components to the implementation, a supervised training pipeline for depth estimation and semantic segmentation, losses (\ref{eq: smticloss}) (\ref{eq: omreg}) (\ref{eq: depthprior}) (\ref{eq: smticprior}), the support for all the datasets used in Section \ref{sect: results}, and the evaluation pipeline. The semantic segmentation network is directly imported from the public implementation of AdapNet++ \cite{valada2019self}. In the self-supervised training phase, dropout \cite{srivastava2014dropout} in AdapNet++ is disabled so that $\hat{S}$ has the best possible quality in the consistency loss (\ref{eq: smticloss}). Inverse depth instead of depth is used throughout the implementation \cite{civera2008inverse}. A simple data augmentation technique of randomly scaling and cropping the input images is employed. Training was done on an NVIDIA TITAN-RTX GPU. 

\section{Results}
\label{sect: results}

We performed experiments on both indoor and outdoor datasets. The indoor datasets, ScanNet \cite{dai2017scannet} and SUN RGB-D \cite{song2015sun}, are recorded in different environments but share many common object categories. CityScapes \cite{Cordts2016Cityscapes} and KITTI \cite{geiger2013vision} are both outdoor driving datasets. All the four datasets consist of real-world images. For semantic segmentation, we used the public checkpoints of AdapNet++ trained on ScanNet, SUN RGB-D, and CityScapes to represent the state-of-the-art supervised training performance on these datasets. Working upon these checkpoints, we performed bootstrapped self-supervised training on the same or different datasets to form the following four combinations, ScanNet to ScanNet, SUN RGB-D to ScanNet, CityScapes to CityScapes, and CityScapes to KITTI. The cases of SUN RGB-D to ScanNet and CityScapes to KITTI fall into the category of unsupervised domain adaptation since the supervised training dataset is different from the self-supervised training dataset. Our evaluation was always performed on the dataset for the self-supervised training in each case. For depth estimation, supervised training was performed on either ScanNet or CityScapes and self-supervised training was on the same datasets as for semantic segmentation. The image resolution was set to $768 \times 384$ to be consistent with the AdapNet++ public checkpoints.

\subsection{ScanNet to ScanNet}
\label{sect: sn2sn}
We performed supervised training on the ScanNet data \cite{dai2017scannet} and then bootstrapped self-supervised training also on ScanNet. Specifically, the supervised training of the AdapNet++ public checkpoint and our depth network were both done on the training split of the ScanNet v2 25k pre-processed frames. In this training set, only the $x00^{th}$ (multiples of 100) frames in each sequence are included, so the supervised training did not take all the frames in the sequences. In the self-supervised training stage, we sampled a 3-frame snippet every 100 frames from the full training sequences with a skip of 10 frames between the successive frames in the snippet. This skip allows enough baseline between the frames in a snippet, since the camera motion in ScanNet is small. The training set we made in this way has only a sparse set of the frames in the original sequences. We could sample more densely, but that would lead to too much data and too much training time. Neither the supervised training nor the self-supervised training used any frames in the validation or test sequences.

All the training and evaluation were done on a set of 20 semantic classes out of the 40 classes available in the ground truth as suggested by the ScanNet benchmark \cite{dai2017scannet}. We used the evaluation code that comes with the ScanNet scripts to evaluate on the full validation split of the 25k pre-processed frames. The results are in Table \ref{tab: sn2sn}.

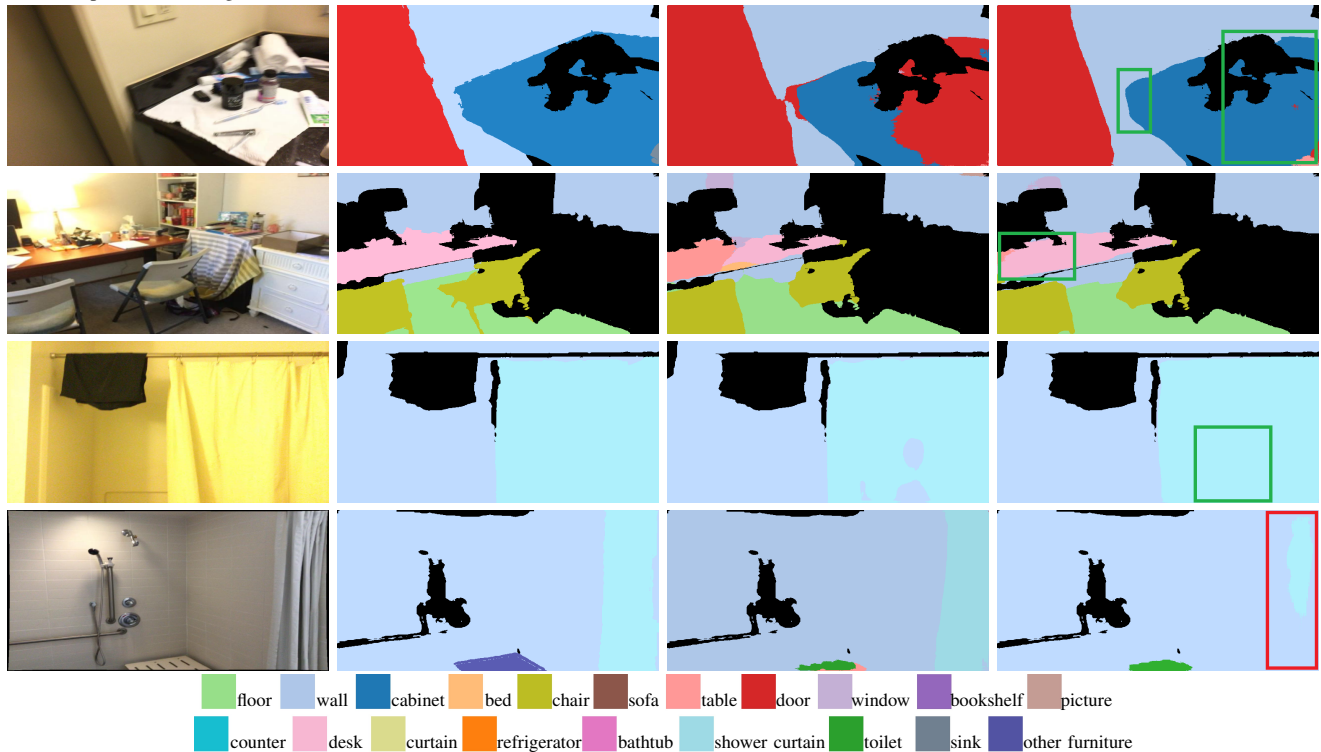
\begin{figure*}[ht]
    \centering
    \begin{subfigure}{0.24\linewidth}
        \includegraphics[width=1.0\columnwidth]{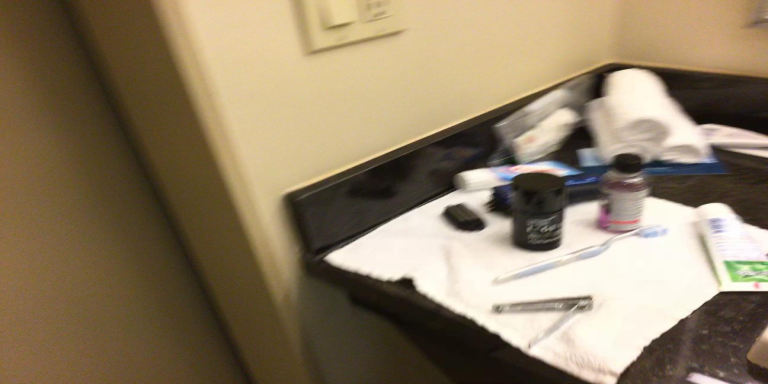}
    \end{subfigure}
    \begin{subfigure}{0.24\linewidth}
        \includegraphics[width=1.0\columnwidth]{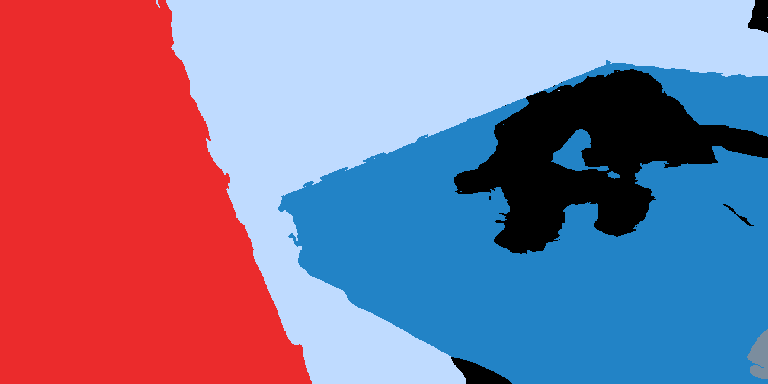}
    \end{subfigure}
    \begin{subfigure}{0.24\linewidth}
        \includegraphics[width=1.0\columnwidth]{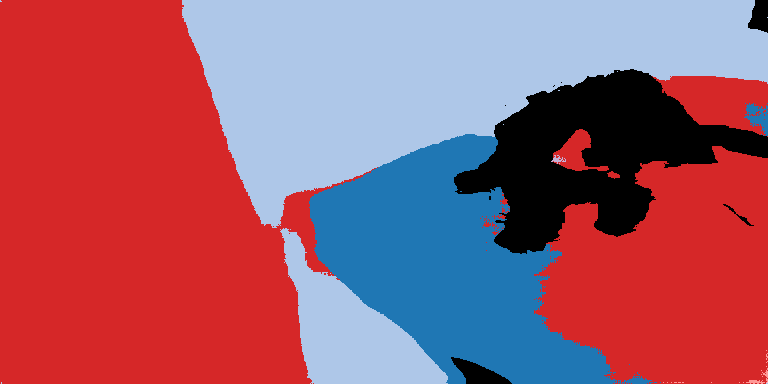}
    \end{subfigure}
    \begin{subfigure}{0.24\linewidth}
        \includegraphics[width=1.0\columnwidth]{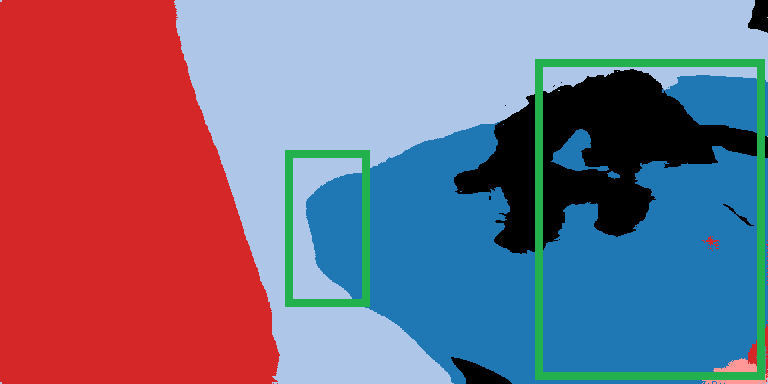}
    \end{subfigure}
    \par\smallskip
    \begin{subfigure}{0.24\linewidth}
        \includegraphics[width=1.0\columnwidth]{figures/img_results/sn_final/img/02304.png}
    \end{subfigure}
    \begin{subfigure}{0.24\linewidth}
        \includegraphics[width=1.0\columnwidth]{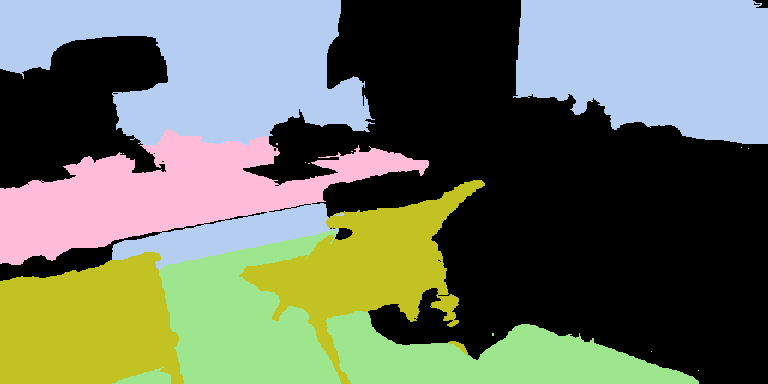}
    \end{subfigure}
    \begin{subfigure}{0.24\linewidth}
        \includegraphics[width=1.0\columnwidth]{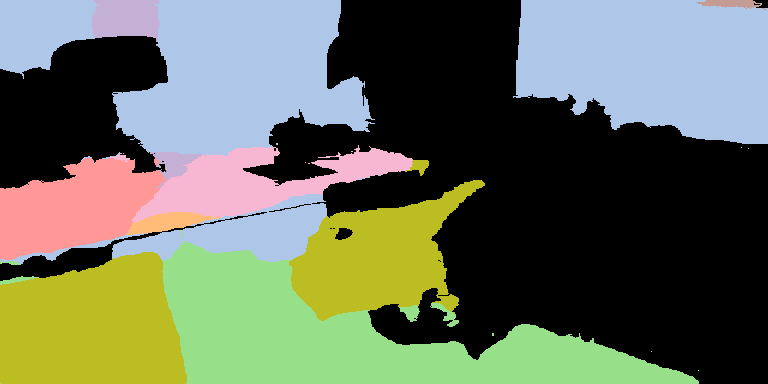}
    \end{subfigure}
    \begin{subfigure}{0.24\linewidth}
        \includegraphics[width=1.0\columnwidth]{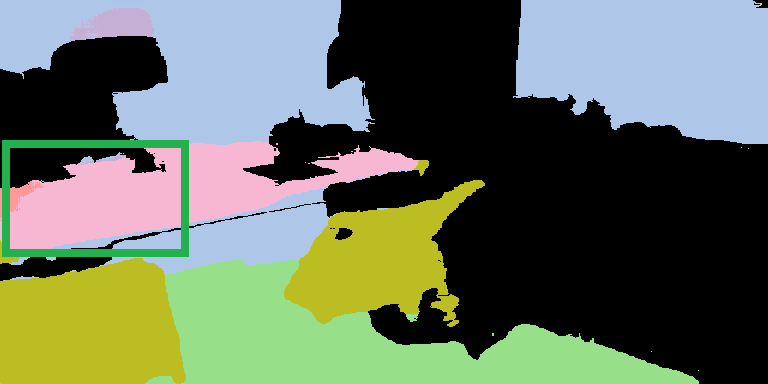}
    \end{subfigure}
    \par\smallskip
    \begin{subfigure}{0.24\linewidth}
        \includegraphics[width=1.0\columnwidth]{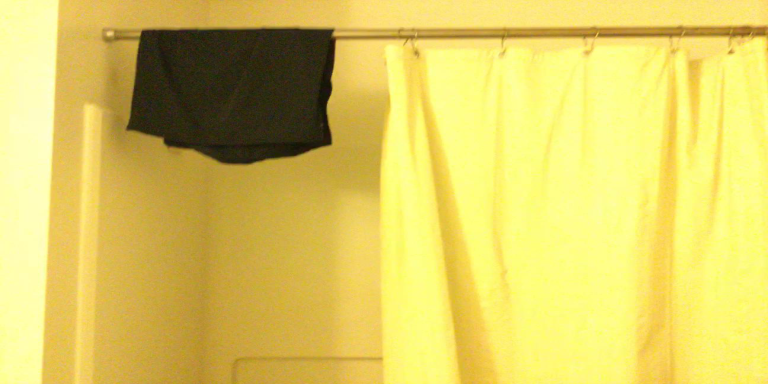}
    \end{subfigure}
    \begin{subfigure}{0.24\linewidth}
        \includegraphics[width=1.0\columnwidth]{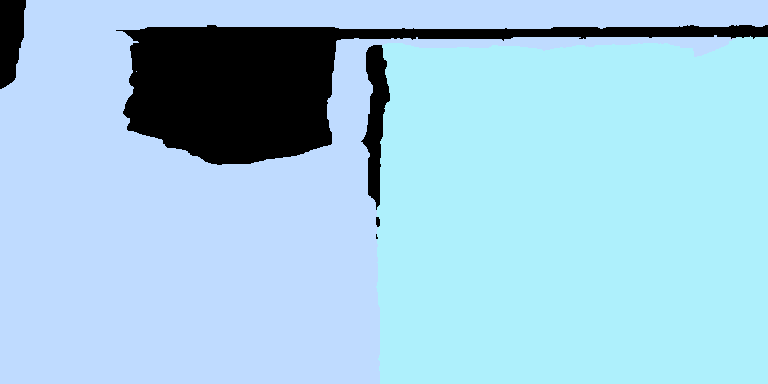}
    \end{subfigure}
    \begin{subfigure}{0.24\linewidth}
        \includegraphics[width=1.0\columnwidth]{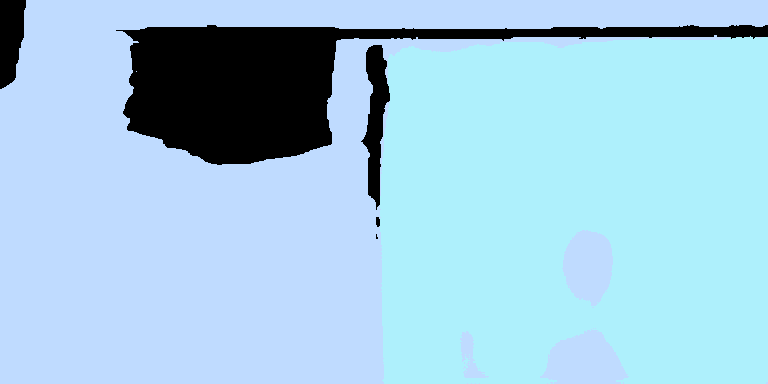}
    \end{subfigure}
    \begin{subfigure}{0.24\linewidth}
        \includegraphics[width=1.0\columnwidth]{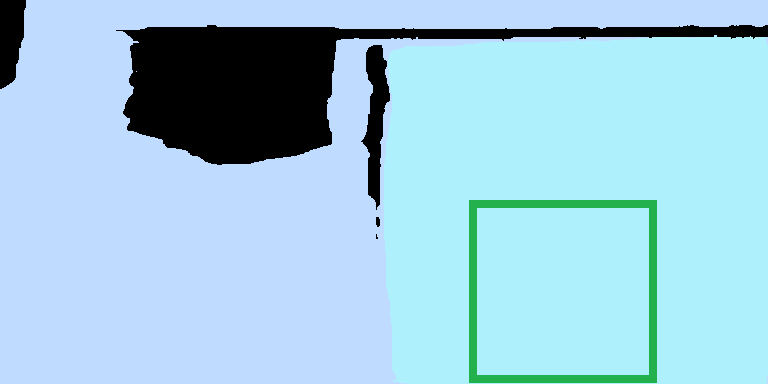}
    \end{subfigure}
    \par\smallskip
    \begin{subfigure}{0.24\linewidth}
        \includegraphics[width=1.0\columnwidth]{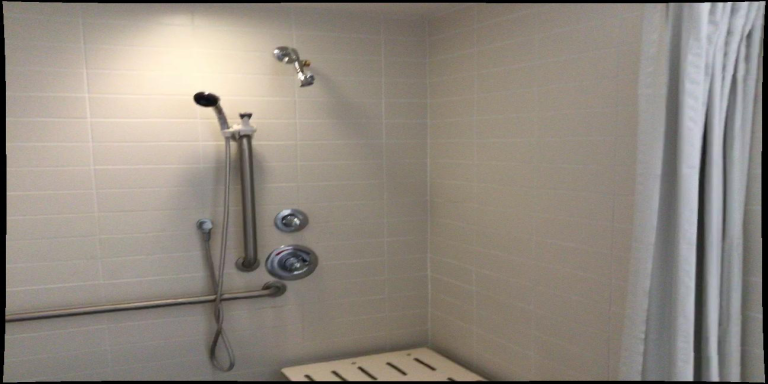}
    \end{subfigure}
    \begin{subfigure}{0.24\linewidth}
        \includegraphics[width=1.0\columnwidth]{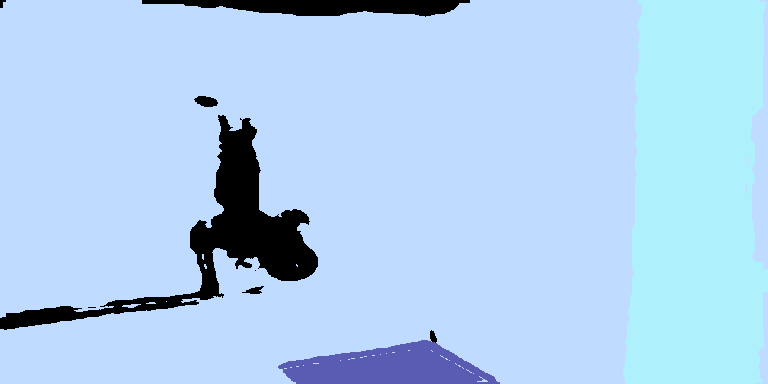}
    \end{subfigure}
    \begin{subfigure}{0.24\linewidth}
        \includegraphics[width=1.0\columnwidth]{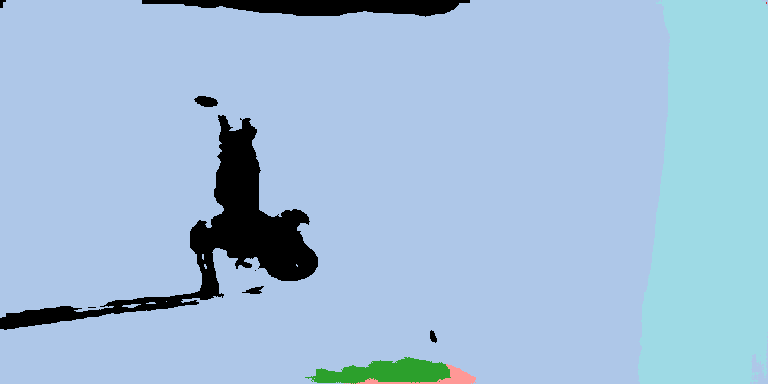}
    \end{subfigure}
    \begin{subfigure}{0.24\linewidth}
        \includegraphics[width=1.0\columnwidth]{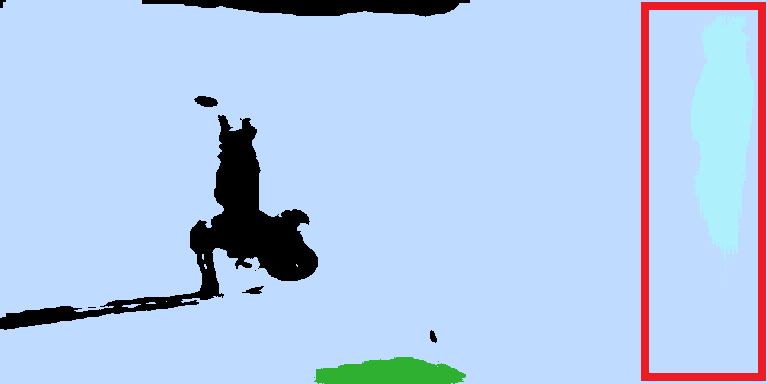}
    \end{subfigure}
    \par\smallskip
    \begin{subfigure}{0.7\linewidth}
        \centering
        \begin{overpic}[width=1.0\columnwidth, tics=2]{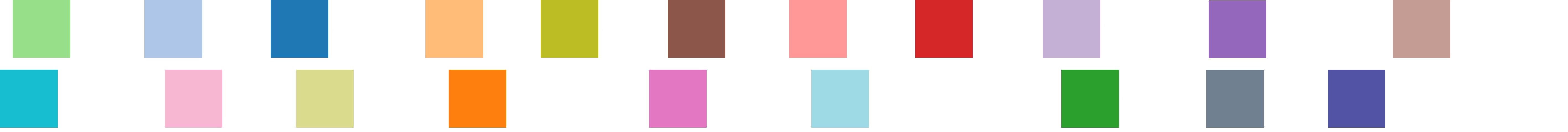}
            \put(4.5,5){\scriptsize floor} \put(13,5){\scriptsize wall} \put(21,5){\scriptsize cabinet}
            \put(31,5){\scriptsize bed} \put(38.2,5){\scriptsize chair}
            \put(46.3,5){\scriptsize sofa} \put(54.1,5){\scriptsize table} \put(62.1,5){\scriptsize door}
            \put(70.1,5){\scriptsize window} \put(80.7,5){\scriptsize bookshelf}
            \put(92.5,5){\scriptsize picture} \put(3.8,0.5){\scriptsize counter}
            \put(14.3,0.5){\scriptsize desk} \put(22.6,0.5){\scriptsize curtain}
            \put(32.3,0.5){\scriptsize refrigerator} \put(45.2,0.5){\scriptsize bathtub} \put(55.5,0.5){\scriptsize shower curtain}
            \put(71.5,0.5){\scriptsize toilet} \put(80.7,0.5){\scriptsize sink} \put(88.5,0.5){\scriptsize other furniture}
        \end{overpic}
    \end{subfigure}
\caption{Example semantic segmentation images before and after the bootstrapped self-supervised training from the ScanNet-to-ScanNet experiment (Section \ref{sect: sn2sn}). From left to right: input image, ground truth semantic segmentation, segmentation produced by the supervised training baseline, segmentation produced by the bootstrapped self-supervised training. Green boxes indicate improved semantic segmentation, and red boxes indicate degraded regions.}
\label{fig: eg_sn2sn}
\end{figure*}

In Table \ref{tab: sn2sn}, the bootstrapped self-supervised training improves the network performance in all classes except the shower curtain class and two others. Shower curtains are easily blended into the background due to color similarity (Figure \ref{fig: eg_sn2sn} rows 3 and 4). The network has the tendency to make mistakes on this class. During training, the semantic consistency loss (\ref{eq: smticloss}) adversely pushed the predictions into consistent mistakes, causing the performance on this class to degrade. However, the performance loss due to those degenerated classes was outrun by the positive improvements in other classes. Figures \ref{fig: intro} and \ref{fig: eg_sn2sn} show some of these positive cases. Overall, the 2.7\% improvement is significant because usually the top competitors on the ScanNet benchmark are only different by a few percent.


\begin{table*}[t]
\par\smallskip 
\par\smallskip
    \flushleft
    \begin{tabular}{|l|l|l| c c c c c c c c c c c c c|}
        \hline
        Sup. & Self. & Test & Mean & Wall & Floor & Cabinet & Bed & Chair & Sofa & Table & Door & Window & Bookshelf & Picture & Counter \\
        \hline
        sn & none & sn & 53.0 & 71.2 & 80.6 & 46.7 & 61.0 & 59.0 & 55.2 & 58.6 & 54.9 & 37.5 & 52.4 & 26.7 & 33.1 \\
        sn & sn & sn & +2.7 & +2.4 & +0.7 & +6.4 & +3.2 & +3.2 & +4.3 & +1.1 & +2.3 & +4.1 & +3.6 & -0.9 & +2.5 \\
        \hline
        sun & none & sn & 39.6 & 67.3 & 74.3 & 28.8 & 44.3 & 52.2 & 45.5 & 45.5 & 38.9 & 30.1 & 16.6 & 28.5 & 23.6 \\
        sun & sn & sn & +2.4 & +1.1 & +2.6 & +5.3 & +3.0 & +2.1 & -0.4 & +4.8 & +4.0 & +2.0 & +13.1 & +6.9 & -5.4 \\
        \hline
    \end{tabular}
    \par\smallskip
    \begin{tabular}{|l|l|l| c c c c c c c c|}
        \hline
        Sup. & Self. & Test & Desk & Curtain & Refrig. & Shower Curtain & Toilet & Sink & Bathtub & Oth. \\
        \hline
        sn & none & sn & 40.0 & 37.6 & 54.6 & 54.9 & 81.7 & 55.1 & 68.7 & 31.6 \\
        sn & sn & sn   & +8.4 & +2.3 & +3.5 & -5.4 & +3.4 & -0.2 & +5.9 & +3.0 \\
        \hline
        sun & none & sn & 21.3 & 30.9 & 44.0 & 1.9 & 66.8 & 43.6 & 49.0 & n/a \\
        sun & sn & sn   & +9.7 & +2.8 & -7.4 & -1.9 & +7.0 & +3.7 & -8.5 & n/a \\
        \hline
    \end{tabular}
    \caption{Class IoU results (\%). The first three columns represent the datasets for the supervised training baseline, bootstrapped self-supervised training upon the baseline, and testing (sn: ScanNet \cite{dai2017scannet}, sun: SUN RGB-D \cite{song2015sun}). We report the improvements that the bootstrapped self-supervised training made over the supervised training.}
    \label{tab: sn2sn}
\end{table*}

\subsection{SUN RGB-D to ScanNet}
\label{sect: sun2sn}
To examine the performance of our method when the supervised training dataset is different from the self-supervised training dataset, we used the AdapNet++ checkpoint for SUN RGB-D \cite{song2015sun} as the baseline. The self-supervised training was performed with the same 3-frame snippets sampled from ScanNet as described in Section \ref{sect: sn2sn}. The AdapNet++ checkpoint for SUN RGB-D was trained on 37 classes which include 19 classes out of the 20 classes in the ScanNet evaluation. We performed the self-supervised training on the full 37 classes, but we evaluated only on the 19 classes. All the other classes were mapped to an additional void class during the evaluation. The results are shown in Table \ref{tab: sn2sn}.

An improvement that is comparable to the previous ScanNet-to-ScanNet result can be seen in this cross-dataset experiment. As in the previous experiment, the degraded classes (e.g. refrigerator and bathtub) in this experiment also often have a white texture which tends to be confused with the background. The semantic consistency aggravated this tendency to make mistakes during training. However, the overall enhancement of 2.4\% demonstrates that our method is still valid for cross-dataset training.

\subsection{CityScapes to CityScapes}
\label{sect: cs2cs}

\begin{table*}[ht]
\par\smallskip
\par\smallskip
    \centering
    \begin{tabular}{|l|l|l| c c c c c c c c c c c c|}
        \hline
        Sup. & Self. & Test & Mean & Sky & Bldg. & Road & Sidewalk & Fence & Vegetation & Pole & Car & Traffic Sign & Pedestrians & Rider\\
        \hline
        cs & none & cs & 80.2 & 94.0 & 91.5 & 96.6 & 83.6 & 54.9 & 92.1 & 56.7 & 93.5 & 72.8 & 74.7 & 72.4 \\
        cs & cs & cs & +0.3 & -0.1 & +0.1 & 0.0 & +0.1 & +1.6 & 0.0 & +0.6 & +0.1 & +0.7 & +0.1 & +0.1 \\
        \hline
        cs & none & k & 53.5 & 92.3 & 77.8 & 86.9 & 40.0 & 28.9 & 87.5 & 42.0 & 73.7 & 39.9 & 12.2 & 6.9 \\
        cs & k & k & +4.2 & +0.7 & +3.1 & +1.7 & +2.7 & +4.1 & +2.4 & +4.7 & +5.9 & +8.5 & +10.0 & +2.9 \\
        \hline
    \end{tabular}
    \caption{Class IoU results (\%). The first three columns represent the datasets for the supervised training baseline, bootstrapped self-supervised training upon the baseline, and testing (cs: CityScapes \cite{Cordts2016Cityscapes}, k: KITTI \cite{geiger2013vision}). We report the improvements that the bootstrapped self-supervised training made over the supervised training. The "car" class also includes trucks and buses. The "rider" class also includes bicycles and motorcycles. For CityScapes, we tested on images with the car logo cropped so the numbers are slightly different from those given by Adapnet++ \cite{valada2019self}.}
    \label{tab: cs2cs}
\end{table*}

\begin{figure*}[ht]
    \centering
    \vspace{12pt}
    \begin{subfigure}{0.24\linewidth}
        \includegraphics[width=1.0\columnwidth]{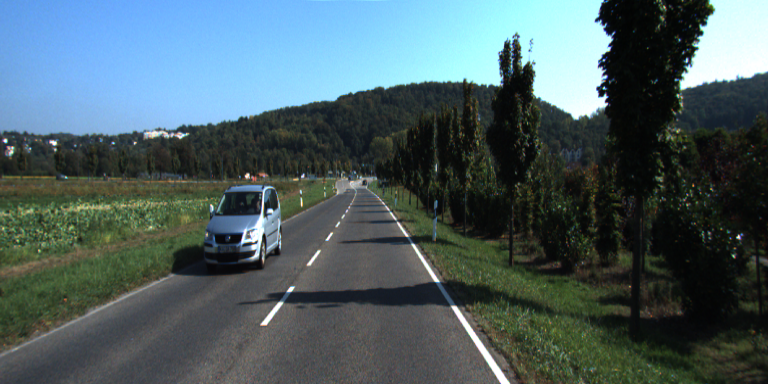}
    \end{subfigure}
    \begin{subfigure}{0.24\linewidth}
        \includegraphics[width=1.0\columnwidth]{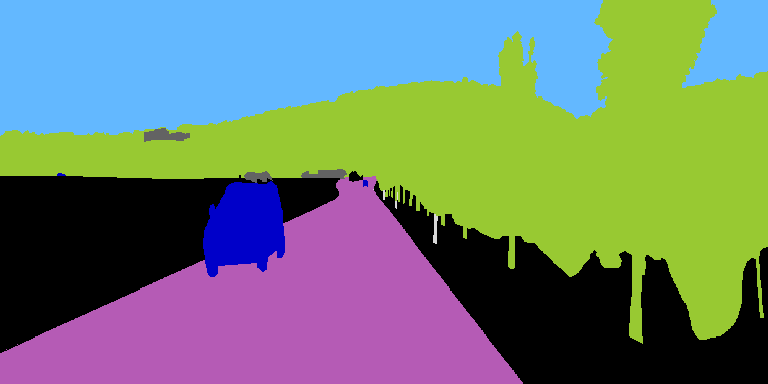}
    \end{subfigure}
    \begin{subfigure}{0.24\linewidth}
        \includegraphics[width=1.0\columnwidth]{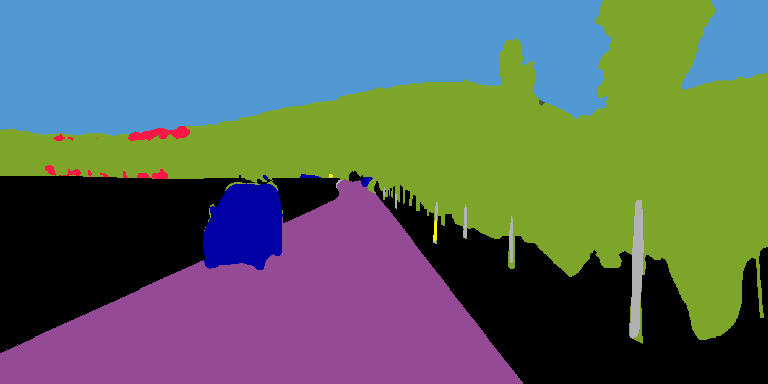}
    \end{subfigure}
    \begin{subfigure}{0.24\linewidth}
        \includegraphics[width=1.0\columnwidth]{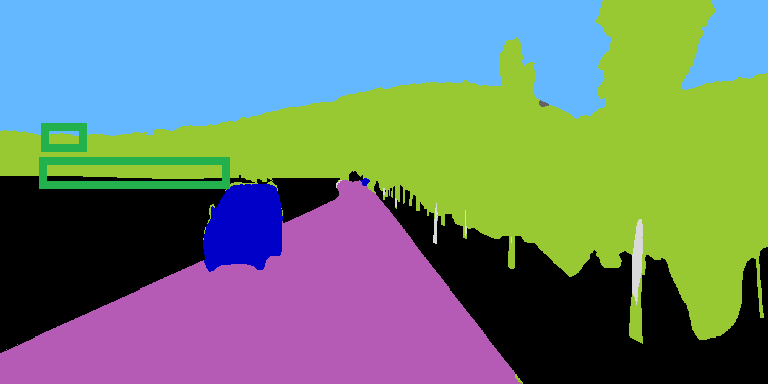}
    \end{subfigure}
    \par\smallskip
    \begin{subfigure}{0.24\linewidth}
        \includegraphics[width=1.0\columnwidth]{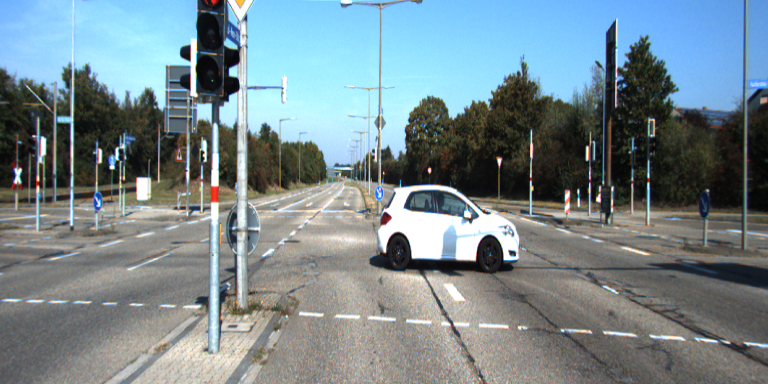}
    \end{subfigure}
    \begin{subfigure}{0.24\linewidth}
        \includegraphics[width=1.0\columnwidth]{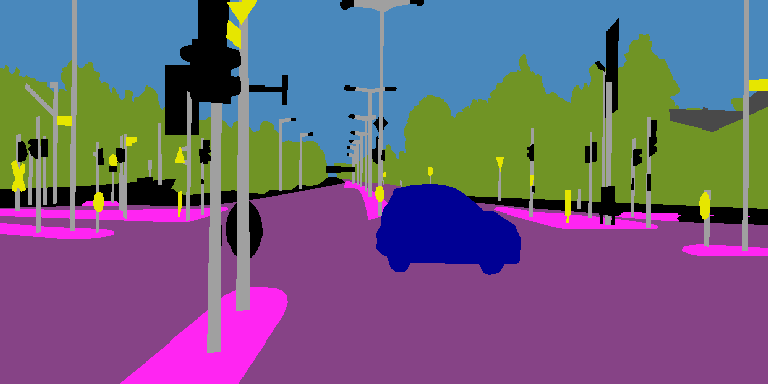}
    \end{subfigure}
    \begin{subfigure}{0.24\linewidth}
        \includegraphics[width=1.0\columnwidth]{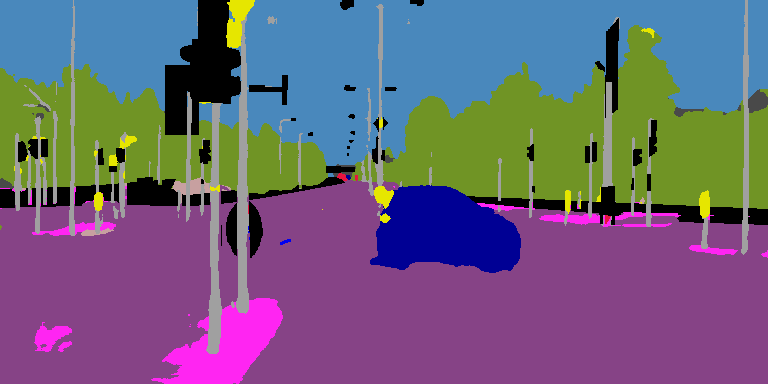}
    \end{subfigure}
    \begin{subfigure}{0.24\linewidth}
        \includegraphics[width=1.0\columnwidth]{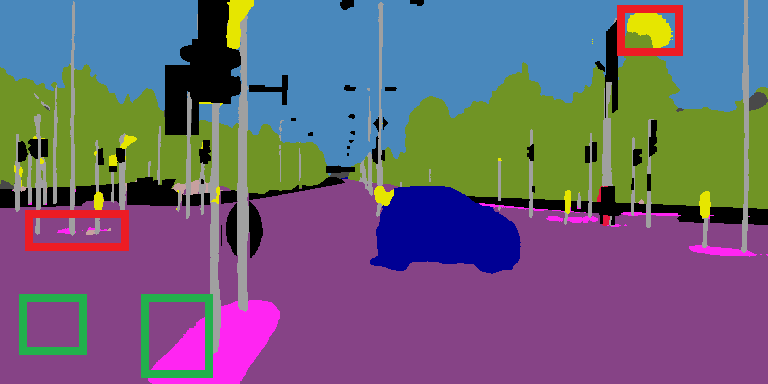}
    \end{subfigure}
    \par\smallskip
    \begin{subfigure}{0.24\linewidth}
        \includegraphics[width=1.0\columnwidth]{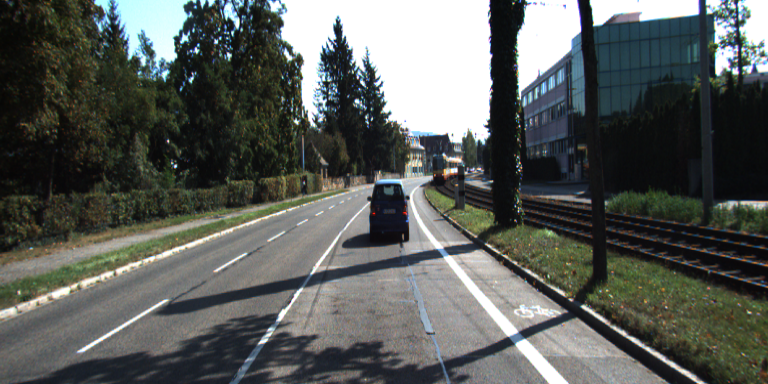}
    \end{subfigure}
    \begin{subfigure}{0.24\linewidth}
        \includegraphics[width=1.0\columnwidth]{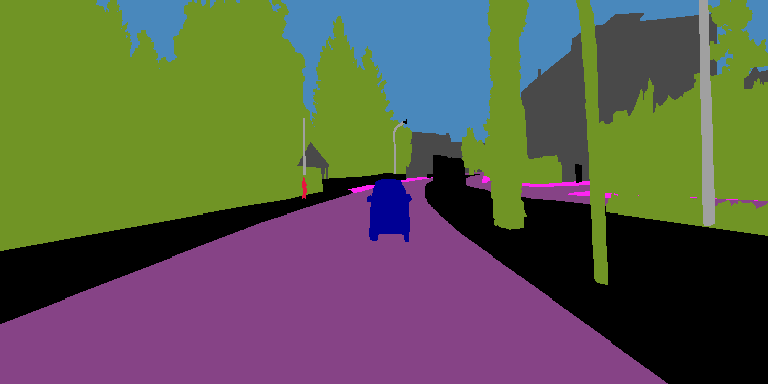}
    \end{subfigure}
    \begin{subfigure}{0.24\linewidth}
        \includegraphics[width=1.0\columnwidth]{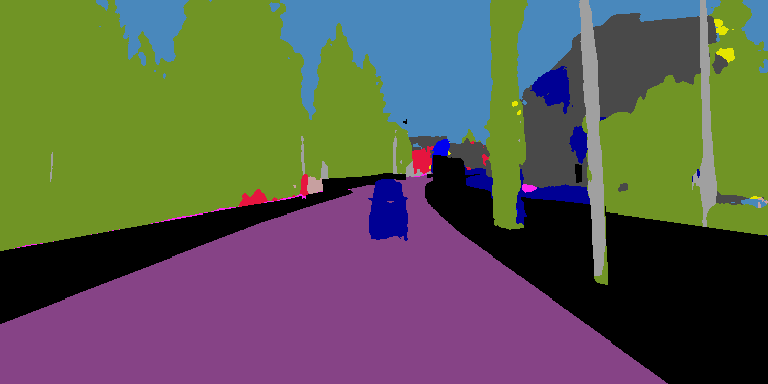}
    \end{subfigure}
    \begin{subfigure}{0.24\linewidth}
        \includegraphics[width=1.0\columnwidth]{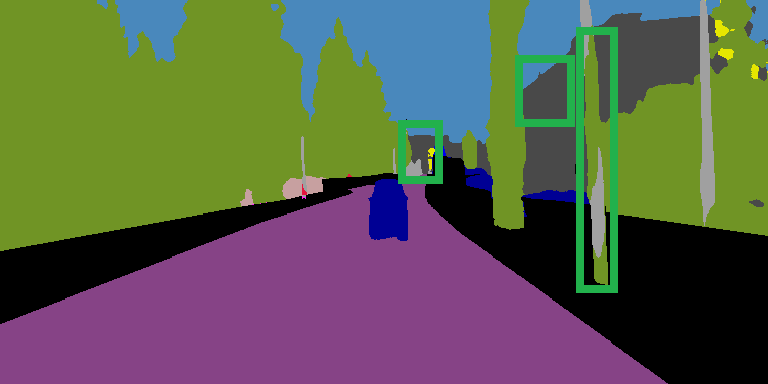}
    \end{subfigure}
    
\caption{Example semantic segmentation images before and after the bootstrapped self-supervised training from the CityScapes-to-KITTI experiment (Section \ref{sect: cs2k}). From left to right: input image, ground truth semantic segmentation, segmentation produced by the supervised training, segmentation produced by the bootstrapped self-supervised training. Green boxes indicate improved semantic segmentation, and red boxes indicate degraded regions.} 
\label{fig: cs}
\end{figure*}

\begin{table*}[h!t]
    \vspace{7pt}
    \centering
    \begin{tabular}{l c c | c c c c | c c c | c}
        \hline
        Method & Sup. & Self. & Abs Rel & Sq Rel & RMSE & RMSE log & $\delta < 1.25$ & $\delta < 1.25^2$ & $\delta < 1.25^3$ & Scale \\
        \hline
        Eigen et al. \cite{eigen2014depth} & k & & 0.203 & \bf 1.548 & 6.307 & 0.282 & 0.702 & 0.890 & 0.958 & n/a \\
        Liu et al. \cite{liu2015learning} & k & & 0.202 & 1.614 & 6.523 & 0.275 & 0.678 & 0.895 & 0.965 & n/a \\
        Zhou et al. \cite{zhou2017unsupervised} & & cs+k & 0.198 & 1.836 & 6.565 & 0.275 & 0.718 & 0.901 & 0.960 & n/a \\
        Ours (416 $\times$ 128) & cs & & 0.200 & 2.138 & 7.085 & 0.267 & 0.734 & 0.909 & 0.962 & \bf 1.02 \\
        Ours (416 $\times$ 128) & & k & 0.181 & 1.760 & 6.373 & 0.260 & 0.751 & 0.911 & 0.962 & 1.09 \\
        Ours (416 $\times$ 128) & cs & k & \bf 0.173 & 1.751 & \bf 6.204 & \bf 0.254 & \bf 0.769 & \bf 0.922 & \bf 0.966 & \bf 0.98 \\
        \hline
        Ours (768 $\times$ 384) & sn & & 0.159 & 0.108 & 0.366 & 0.219 & 0.810 & 0.943 & 0.978 & \bf 0.97 \\
        Ours (768 $\times$ 384) & & sn & 0.626 & 3.610 & 1.655 & 0.498 & 0.538 & 0.771 & 0.865 & 6.83 \\
        Ours (768 $\times$ 384) & sn & sn & \bf 0.156 & \bf 0.105 & \bf 0.359 & \bf 0.215 & \bf 0.820 & \bf 0.947 & \bf 0.980 & \bf 0.97 \\
        \hline
    \end{tabular}
    \vspace{3pt}
    \caption{Depth prediction evaluation on Eigen et al.'s test split \cite{eigen2014depth}. The abbreviations, sup. and self., indicate on what dataset(s) the supervised training and self-supervised training were performed (cs: CityScapes \cite{Cordts2016Cityscapes}, k: KITTI \cite{geiger2013vision}, sn: ScanNet \cite{dai2017scannet}). $\delta$ means the ratio between the predicted depth and the ground truth depth. The scale factors were obtained through the evaluation code from \cite{zhou2017unsupervised}. A same constant scale factor was pre-multiplied (before training) to offset the depth scale difference between CityScapes and KITTI for the supervised training and bootstrapped self-supervised training cases.}
    \label{tab: k_depth}
\end{table*}

We further investigated our approach on an outdoor dataset, CityScapes \cite{Cordts2016Cityscapes}. AdapNet++ has two public checkpoints for CityScapes, one with 19 classes and one with 11 classes. The 19-class checkpoint uses different image resolutions for training and testing, rendering it unsuitable to integrate into our method, so we chose the 11-class checkpoint. The checkpoint and our depth network were both trained on the training split of the 5000 finely annotated images in CityScapes. The same data preparation procedures as in \cite{mahjourian2018unsupervised} were engaged to generate the 3-frame snippets. We additionally cropped the car logo in the images for both the supervised training and the bootstrapped self-supervised training to be consistent with the KITTI data \cite{geiger2013vision} (used in the CityScapes-to-KITTI experiment in Section \ref{sect: cs2k}). The evaluation was performed on those 11 classes included in the AdapNet++ checkpoint and on images with the car logo cropped. The results are in Table \ref{tab: cs2cs}.

CityScapes is a small dataset with only 2975 training images and the baseline model is trained on a subset of 11 classes. Thus, the high performance of the baseline may be partially attributed to overfitting on the small amount of data. However, our method still manages to make a small overall improvement over the baseline. The improvement is lower than 1.08\% in \cite{zhu2019improving} which is a pseudo-labeling approach with video propagation techniques. This can be due to our bootstrapped self-supervised training scenario being more challenging than the pseudo-labeling approach as discussed in Section \ref{sect: literature}.



\subsection{CityScapes to KITTI}
\label{sect: cs2k}

We performed self-supervised training on KITTI \cite{geiger2013vision} bootstrapping from the models trained on CityScapes \cite{Cordts2016Cityscapes}. We followed \cite{mahjourian2018unsupervised} to prepare the 3-frame snippets data from the training set of the KITTI raw data \cite{geiger2013vision} according to Eigen et al.'s split \cite{eigen2014depth}. The KITTI odometry data \cite{geiger2012we} were not used. We evaluated on the KITTI semantics 200 training images \cite{geiger2013vision}. The results are given in Table \ref{tab: cs2cs} and Figure \ref{fig: cs}. 

In Figure \ref{fig: cs}, a common mistake that the baseline model makes is that far plants and buildings are often mis-classified as pedestrians or traffic signs. This is possibly caused by the domain gap between CityScapes \cite{Cordts2016Cityscapes} and KITTI \cite{geiger2013vision} since CityScapes more often contains busy city street views with lots of pedestrians while the KITTI environment is more rural. Our bootstrapped self-supervised training approach greatly suppresses these false positives as shown in Figure \ref{fig: cs} and Table \ref{tab: cs2cs}. This further demonstrates our approach as an effective strategy for domain adaptation.



\subsection{Depth and Ego-Motion Estimation Evaluation}

We evaluated the depth predictions from our network on KITTI and ScanNet. These two datasets were selected because KITTI is a widely used dataset for the depth estimation evaluation and ScanNet is an indoor dataset where pure self-supervised learning methods for depth and ego-motion do not yield high-quality results due to the jittering and full 6-degree-of-freedom motion of the hand-held camera.

For KITTI, we tested the depth estimation on the Eigen et al.'s split \cite{eigen2014depth} using the public evaluation code from \cite{zhou2017unsupervised} which computes a scale factor first to cancel the scale difference between the predicted depth and the ground truth, and then computes several metrics. In Table \ref{tab: k_depth}, we compare our bootstrapped self-supervised training method to pure supervised approaches \cite{eigen2014depth, liu2015learning} and a pure self-supervised approach \cite{zhou2017unsupervised}. We compare to \cite{zhou2017unsupervised} because their losses to train the depth network are very similar to ours, so the main difference is the bootstrapping part. Additionally, we compare our results of supervised-only, self-supervised-only, and bootstrapped self-supervised training. The supervised-only results reflect the model performance before our bootstrapped self-supervised training. The self-supervised-only results were obtained by training the depth network fresh using losses (\ref{eq: photoloss}), (\ref{eq: ssimloss}), (\ref{eq: dsmth}), (\ref{eq: omreg}). The hyperparameters were set same as in the bootstrapped self-supervised training. We train our model on $416 \times 128$ image resolution to be consistent with \cite{zhou2017unsupervised}. The same ablation study was also performed on ScanNet. The evaluation was done on the validation set of the ScanNet v2 25k pre-processed frames.

As seen in Table \ref{tab: k_depth}, the bootstrapped self-supervised training performance is superior to the pure supervised or self-supervised training in terms of prediction accuracy. In particular, by incorporating the prior loss (\ref{eq: depthprior}), it successfully learned depth estimation from video recorded by a hand-held camera (ScanNet), which is quite difficult for pure self-supervised training as indicated in Table \ref{tab: k_depth}. In addition, unlike the pure self-supervised training results which are drawn to arbitrary depth scales, the bootstrapped self-supervised training can retain the absolute scale of the predicted depth from the supervised training.

Although the ego-motion network is auxiliary in this work, we evaluate the trained ego-motion network in Table \ref{tab: k_odom} for completeness using the evaluation code from \cite{zhou2017unsupervised}. The network was trained along with the depth network in the depth estimation experiments on KITTI. As mentioned in Section \ref{sect: nets}, unlike the depth network, the ego-motion network does not undergo the supervised training stage. The training signal to the ego-motion network is only from the self-supervision with video. We found that the ego-motion prediction results were the same (up to the third decimal place) both when the depth network was trained in the pure self-supervised fashion and when the depth network was trained in the bootstrapped self-supervised fashion. This finding suggests that the improvement in the depth estimation by doing the bootstrapped self-supervised training (Table \ref{tab: k_depth}) was not transmitted to the ego-motion prediction in our experiments.

\begin{table}
    \centering
    \begin{tabular}{l c c}
        \hline
        Method & Seq. 09 & Seq. 10 \\
        \hline
        Zhou et al. \cite{zhou2017unsupervised} & 0.021 $\pm$ 0.017 & 0.020 $\pm$ 0.015 \\
        Ours (416 $\times$ 128) & \bf 0.014 $\pm$ 0.012 & \bf 0.011 $\pm$ 0.013 \\
        \hline
    \end{tabular}
    \vspace{3pt}
    \caption{Ego-motion prediction evaluated on the KITTI odometry sequences \cite{geiger2012we}. Absolute trajectory error (ATE) averaged over the multi-frame snippets is used as the evaluation metric.}
    \label{tab: k_odom}
\end{table}


\section{Conclusion}
We have demonstrated the effectiveness of our bootstrapped self-supervised training method in improving semantic segmentation over a supervised training baseline on both indoor and outdoor datasets. We have also shown that our method can increase the accuracy in depth estimation and retain the absolute depth scale in the prediction. With our bootstrapped self-supervised training method, we imagine a robot only needs to train on labeled data initially and can perform fully self-supervised training afterwards to improve itself and adapt to new environments.









\bibliographystyle{IEEEtran}
\bibliography{IROS2021_zyh}

\end{document}